\documentclass{article}
\usepackage[nonatbib, final]{nips_2018}
\usepackage{nips_2018}
\usepackage[utf8]{inputenc} 
\usepackage[T1]{fontenc}    
\usepackage{hyperref, authblk}       
\usepackage{url}            
\usepackage{booktabs}       
\usepackage{amsfonts}       
\usepackage{nicefrac}       
\usepackage{microtype}      
\usepackage{blindtext}
\usepackage{amsmath, psfrag}
\usepackage{amssymb, relsize}
\usepackage{textcomp}
\usepackage{graphicx}
\usepackage[export]{adjustbox}
\usepackage{subcaption,dirtytalk}
\usepackage{tikz}
\usepackage{pgfplots}
\usepackage{setspace}
\usepackage{multirow}
\usepackage{epstopdf,caption}
\usepackage{multirow}
\title{Taxi Demand-Supply Forecasting: Impact of Spatial Partitioning on the Performance of Neural Networks}

\author{\textbf{Neema Davis} \hspace{0.5cm} \textbf{Gaurav Raina} \hspace{0.5cm} \textbf{Krishna Jagannathan}\\
  Department of Electrical Engineering\\
  Indian Institute of Technology Madras  \\
  Chennai, India 600 036 \\
  \texttt{\{ee14d212, gaurav, krishnaj\}@ee.iitm.ac.in} \\
}

\begin{document}


\maketitle

\begin{abstract}
  In this paper, we investigate the significance of choosing an appropriate tessellation strategy for a spatio-temporal taxi demand-supply modeling framework. Our study compares (i) the variable-sized polygon based Voronoi tessellation, and (ii) the fixed-sized grid based Geohash tessellation, using taxi demand-supply GPS data for the cities of Bengaluru, India and New York, USA. Long Short-Term Memory (LSTM) networks are used for modeling and incorporating information from spatial neighbors into the model. 
  We find that the LSTM model based on input features extracted from a variable-sized polygon tessellation yields superior performance over the LSTM model based on fixed-sized grid tessellation. Our study highlights the need to explore multiple spatial partitioning techniques for improving the prediction performance in neural network models.  
\end{abstract}

\section{Introduction}
Intelligent Transportation Systems (ITSs) play a major role in traffic management, incident response, safety enhancement, and resource utilization. In e-hailing taxi services, the balance between passenger demand and driver supply is an important factor in reducing costs and increasing driver utilization. 
In peak hours, the limited supply of taxis cannot match the high customer demand and in off-peak hours, the surplus taxis are met with low demand. This demand-supply mismatch can be mitigated by devising efficient location-based demand-supply forecasting algorithms. In this context, judicious spatial partitioning can play a significant role in enhancing the performance of models.
\subsection{Related works}
The modeling of taxi demand and driver supply have been active areas of research in the transportation literature. Most of the existing works are unilateral; such as \cite{ke2017short,tong2017simpler} which consider only demand forecasting, or \cite{Zhang2015UnderstandingTS,lee2017taxi} which deal with only taxi supply analysis. To develop a comprehensive understanding about the macroeconomic demand-supply patterns in a city, it is essential to perform a combined analysis of demand and supply. 
Spatio-temporal correlations have been shown to enhance the model performance, when integrated into the modeling framework; for example, see \cite{gang2016review, ermagun2018spatiotemporal}. 
Spatial and temporal variants of Auto Regressive models \cite{friedrich2010short,faghih2017predicting}, K-Nearest Neighbor models \cite{cheng2018short}, Support Vector Machines \cite{yao2017short} have been proposed to infuse historic temporal and spatial neighbor information together, resulting in improved prediction accuracy for traffic applications. 
Active research in the area of neural network based taxi travel forecasting has resulted in Residual networks \cite{zhang2017deep}, Convolutional Long Short-Term Memory (LSTM) networks \cite{xingjian2015convolutional}, Attention networks \cite{zhou2018predicting}, to list a few. Even though the network structure varies with each technique, we observe that the input feature information to these networks are usually obtained from a city structure partitioned using fixed-sized grids. That is, whenever location-based forecasting is concerned, it is common practice in the literature to partition the city into fixed-sized grids and capture information from these grids and their neighbors. We notice that LSTM units are widely used as building blocks in the spatio-temporal neural networks \cite{xingjian2015convolutional,zhou2018predicting,yao2018deep}. In this scenario, we believe it is imperative to examine the impact of different partitioning strategies on the prediction performance of the LSTM models. In fact, it might aid the selection of an appropriate tessellation strategy for studies involving spatial correlation. To the best of our knowledge, this aspect has not been studied in the transportation literature.  
\subsection{Our contributions}
We compare the variable-sized polygon based Voronoi tessellation and fixed-sized grid based Geohash tessellation on the demand-supply data sets obtained from two cities, using three widely employed performance metrics. We vary the spatial features to identify the best LSTM configuration. The main findings of this paper are summarized below:
\begin{itemize}
    \item The LSTM network achieves higher prediction accuracy with Voronoi tessellation based input features than with Geohash tessellation based features.
    \item The variability in the prediction performance is lower with Voronoi tessellation based models than with Geohash tessellation based models.
\end{itemize}   

The rest of the paper is organized as follows. In Section \ref{problemsetting}, we describe the problem setting. We explain the modeling methodology in Section \ref{methodology}. Then, we discuss the results in Section \ref{resultssection}, and we outline the conclusions in Section \ref{conclusions}. 

\section{Problem setting} \label{problemsetting}
For location-based forecasting, the area of interest is divided into a set of regions $R$ = $\{r_1,...,r_N\}$. The regions can be grids of fixed or variable-sized area, zip codes, states, \emph{etc}. Once the regions are defined, the data is aggregated per region and modeled for analysis or forecasting applications. Often data from the neighboring regions is included as input features to boost the prediction performance \cite{ermagun2018spatiotemporal}. In this paper, we explore information from the first-order spatial neighbors, \emph{i.e.}, from the neighbors sharing a boundary with the region of interest, as is commonly done in the literature \cite{gang2016review}. Our goal is to observe and analyze the changes in behavior of the spatio-temporal model with the partitioning technique employed. The investigation is performed by modeling the data $d_i$ in each region $r_i$ as:
\begin{equation}
     d_{i,t} = f(d_{i,\{1,\cdots ,t-1\}},d_{\{\forall{j\ \in{\ J}\}},\{1,\cdots,t-1\}}), \text{where}\ J \text{\ is the set of Spatial Neighbors}(i),
\end{equation}
where, $f(\cdot)$ is the non-linear relationship between the variables in a region and its neighbors. We use the LSTM to model this relationship. To obtain a holistic view of the impact of the Voronoi and Geohash tessellations on the predictive capabilities of LSTM, we employ three performance metrics; Symmetric Mean Absolute Percentage Error (SMAPE), Mean Absolute Scaled Error (MASE), and Root Mean Square Error (RMSE). These metrics are defined and discussed in Section \ref{resultssection}.

\section{Modeling methodology}\label{methodology}
We provide a detailed modeling methodology of Bengaluru demand and supply data sets in this section. The same methodology is applied to the New York demand data set and the results observed are presented in Section \ref{nyc}. The primary step in the analysis is to tessellate the city into clusters, for which the K-Means \cite{macqueen1967some} algorithm is employed. These clusters later aid in hot spot identification, and Voronoi cell generation.

\subsection{Data description}\label{data}
The Bengaluru taxi demand and driver supply data sets are acquired from a leading Indian e-hailing taxi service provider. The demand data contains GPS traces of taxi passengers booking a taxi by logging into their mobile application. The supply data contains GPS traces of fresh log-ins of taxi drivers, representing available supply. The data sets are available for a period of two months; from $\text{1}^{st}$ of January 2016 to $\text{29}^{th}$ of February 2016. The data sets contain latitude-longitude coordinates of the customer/driver, session duration and time stamp. The latitude and longitude coordinates of the city are 12.9716° N, 77.5946° E, with an area of approximately 740 $\text{km}^2$. 

\subsection{Tessellation strategies}\label{tessellation}
Since we work with very large data sets, we employ the K-Means algorithm which has a linear memory and time complexity \cite{xu2015comprehensive}. The algorithm classifies data into a fixed number of clusters (K) by minimizing the squared error function given by:
\begin{equation}
    J = \sum_{j=1}^K \sum_{i=1}^n ||y_{(i)}^j - c_j||^2,
\end{equation}
where $||y_{(i)}^j - c_j||^2$ is the Euclidean distance between a data point $y_{(i)}^j$ and its center $c_j$, and $n$ is the total number of data points. In order to ensure efficient routing of taxis, we limit the average size of a cluster to 1 $\text{km}^2$. Hence, the parameter K is set to 740, resulting in 740 clusters for Bengaluru, covering an area of 740 $\text{km}^2$. 

The K-Means is performed on demand data, and the demand centroids are used as generating sites for the demand and supply Voronoi cells. This enables us to analyze both demand and supply for the same demand region. Once the centroids are obtained from the K-Means algorithm, they serve as generating sites for the Centroidal Voronoi tessellation. This is a spatial partitioning method that divides space according to the nearest neighbour-rule. Based on the closeness of sites, this tessellation strategy produces polygon partitions of varying areas. 

Geohash tessellation is an extension of the basic grid based technique, with a naming convention. It encodes a latitude-longitude coordinate into an alphanumeric string. In this paper, Geohash (first G capitalized) refers to the technique, and geohash (all small letters) refers to the resultant string. A 6-level geohash is comprised of 6 characters, covers an area of 1.2 km $\times$ 0.6 km (approximately 1 $\text{km}^2$), and is used in our analysis. We find the 6-level geohashes corresponding to the K-Means centroids. 

\begin{figure}[htb]
\begin{subfigure}{0.385\textwidth}
\begin{tikzpicture}
\begin{axis}[enlargelimits=false, 
                 axis on top,
                 height = 4.15cm,
                 width=\textwidth,
                 ylabel={\footnotesize{Longitude}},
                 xlabel={\footnotesize{Latitude}},
                  ylabel near ticks,
                xlabel near ticks,
                 ytick={0.1,0.9},
                 xticklabels={\scriptsize{12.75},\scriptsize{13.15}},
                 xtick={76.1,76.9},
                 yticklabels={\scriptsize{77.45},\scriptsize{77.8}},
                 yticklabel style={rotate=0}
                  ]
    \addplot graphics [ymin=0,ymax=1,xmin=76,xmax=77]{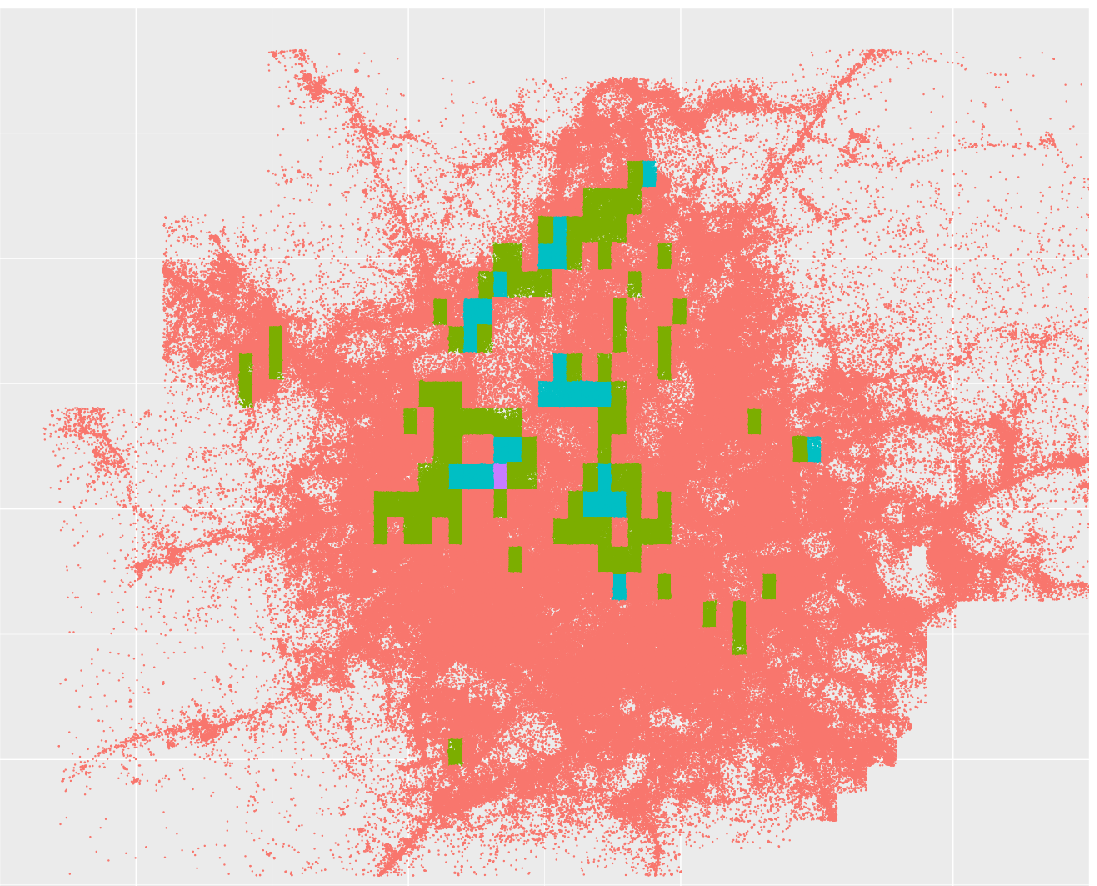};
    \end{axis}
 \end{tikzpicture}
 \vspace{-0.2cm}
  \caption{Geohash tessellation}
\label{voronoi_hm}
\end{subfigure}%
\begin{subfigure}{0.385\textwidth}
\begin{tikzpicture}
\begin{axis}[enlargelimits=false, 
                 axis on top,
                 height = 4.15cm,
                 width=\textwidth,
                 ylabel={\footnotesize{Longitude}},
                 xlabel={\footnotesize{Latitude}},
                  ylabel near ticks,
                xlabel near ticks,
                 ytick={0.1,0.9},
                 xticklabels={\scriptsize{12.75},\scriptsize{13.15}},
                 xtick={76.1,76.9},
                 yticklabels={\scriptsize{77.45},\scriptsize{77.8}},
                 yticklabel style={rotate=0}
                  ]
    \addplot graphics [ymin=0,ymax=1,xmin=76,xmax=77]{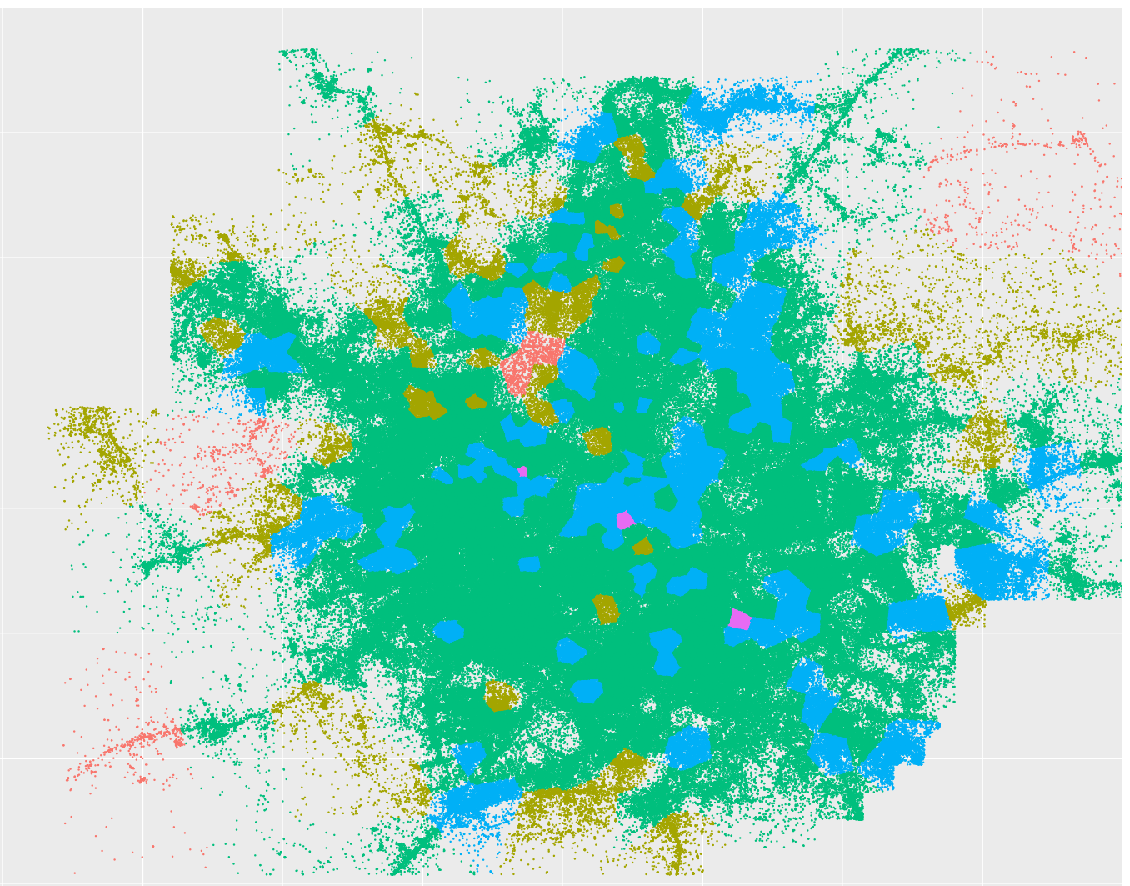};
    \end{axis}
\end{tikzpicture}
\vspace{-0.2cm}
 \caption{Voronoi tessellation}
\label{geohash_hm}
\end{subfigure}%
\begin{subfigure}{0.2\textwidth}
\begin{picture}(100,100)
  \put(15,30)  {\includegraphics[scale=0.175]{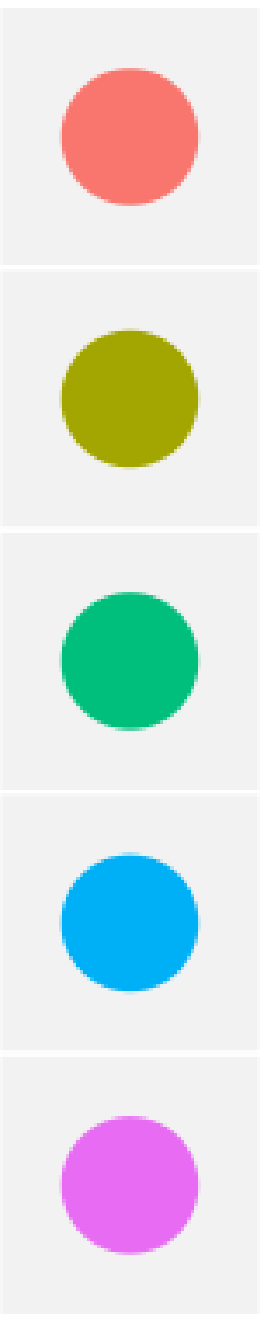}}
  \put(35,87) {\footnotesize{0-5000}}
  \put(35,75) {\footnotesize{5000-10000}}
  \put(35,60) {\footnotesize{10000-25000}}
  \put(35,47) {\footnotesize{25000-50000}}
  \put(35,32) {\footnotesize{50000-150000}}
\end{picture}
\vspace{-1cm}
\caption{Data per cell}
\end{subfigure}%
\caption{Heat maps obtained by partitioning Bengaluru into Voronoi cells and 6-level geohashes.} \label{heatmaps}
\end{figure}

The Voronoi and Geohash heat maps are plotted in Figure \ref{heatmaps}, where the scale denotes the data volume in each cell. Geohash strategy produces rectangular grids of fixed area, resulting in several demand dense and scarce cells. Voronoi strategy tends to uniformly distribute data, and produces polygons of variable area. 

\subsection{Model description}\label{datamodel}
Traditional neural networks lack the ability to learn temporal dependencies, which was overcome by Recurrent Neural Networks (RNN). Long Short-Term Memory (LSTM) networks \cite{lipton2015critical} are variants of RNN, capable of learning long-term temporal dependencies. LSTM appears to be an appropriate choice for our data modeling, as the demand and supply have shown daily and weekly seasonality that are hard to capture using traditional neural networks. For a region $r_i$ (Voronoi cell/geohash), our methodology involves (i) feeding the demand/supply from $r_i$ alone to the LSTM network and (ii) feeding data from $r_i$ and its first-order neighbors to the LSTM network. For each centroid, its corresponding Voronoi cell and 6-level geohash are identified. Then, the demand and supply corresponding to these regions are aggregated every 60 minutes and represented in the form of time-sequences. Each time-sequence contains 1440 time-steps, spanning over 60 days. We use 59 days for training and test it on the $\text{60}^{th}$ day. We also set aside 10\% of training data for validation.   
\subsubsection{Parameter setting}
The performance of deep learning models is dependent on the choice of the hyper-parameters. Bayesian optimization is a preferred technique for optimizing hyper-parameters of various machine learning algorithms \cite{shahriari2016taking}. In this study, we use the Tree-structured Parzen Estimator (TPE) Bayesian optimization approach \cite{bergstra2011algorithms} for tuning the hyper-parameters. This algorithm uses tree-structured Parzen estimators to model the error distribution as non-parametric densities. We ran the TPE optimization over the range of the hyper-parameters outlined in Table \ref{ranges}. Once the parameters are set, the features are fed to the LSTM with the input shape (740, 1440, $s$), where we have 740 regions of interest, 1440 time steps, and $s$ number of spatial features. While a geohash has 8 fixed first-order neighbors, a Voronoi cell can have variable number of neighbors. For consistent comparison, the top 8 positively correlated Voronoi neighbors are considered, where we add invalid vectors to compensate for the lack of features. The number of spatial neighbors are also varied, to explore its impact on the model performance. The LSTM network is trained for 500 epochs with early stopping, and repeated 10 times, keeping in mind the random initialization of network weights. Mean Squared Error (MSE) is chosen as the loss function. 
\vspace{-2mm}
\begin{table}[h!]
  \caption{Range of the hyper-parameters that are fed to the TPE Bayesian optimization algorithm.}
  \label{ranges}
  \centering
  \begin{tabular}{ll}
    \toprule
    Hyper-parameter     & Range    \\
    \midrule
    Number of layers, L & [1, 2] \\
    Number of neurons, n & [10, 20, 50, 100] \\
    Dropout, D     & Uniform (0,0.5) in each layer \\
    Activation     & \{Sigmoid, Relu, Linear\}  \\
    Optimizer     & \{Adam, Stochastic Gradient, RMSprop\}  \\
    Learning Rate     & [$10^{-1}, 10^{-2}, 10^{-3}, 10^{-4}, 10^{-5}, 10^{-6}$]  \\
    Batch size     & [64,128]  \\
    \bottomrule
  \end{tabular}
\end{table}
\vspace{-1mm}
\section{Results}\label{resultssection}
The three error metrics used for the performance evaluation are given below:
\begin{align}
   & 1.\ \text{Symmetric Mean Absolute Percentage Error (SMAPE)} = \frac{100}{N} \mathlarger{\sum_{t=1}^{N}}\frac{|{y}_{t}-\hat{y}_{t}|}{\hat{y}_{t}+y_{t}+1},\\
    & 2.\ \text{Mean Absolute Scaled Error (MASE)} = \frac{1}{N} \mathlarger{\sum_{t= 1}^N}\Bigg (\frac{|y_t - \hat{y_t}|}{\frac{1}{N-m}\sum_{t=m+1}^N |y_t - y_{t-m}|} \Bigg ), \text{and} \\
    & 3.\ \text{Root Mean Square Error (RMSE)} = \sqrt{\frac{1}{N}{\sum_{t=1}^N (y_t - \hat{y_t})^2}}.
\end{align}
where $N$ is the forecast horizon, $m$ is the seasonal period, $y_t$ is the actual demand, and $\hat{y_t}$ is the forecast at time $t$. While the RMSE gives relatively high weights to large errors, the SMAPE is a scale dependent percentage error. The MASE compares the forecast errors from the model with the in-sample forecast errors from the standard Na\"{i}ve model, making it scale independent. The shortlisted LSTM hyper-parameters and the resultant prediction errors are provided in Table \ref{results}. 

\subsection{Bengaluru demand and supply data}
\begin{figure*}[htp!]
\centering
\begin{subfigure}{\textwidth}
\centering
     \psfrag{n}{\hspace{-11mm} \raisebox{-3mm}{\footnotesize{Spatial features}}}
        \psfrag{s1}{\hspace{-5mm}\raisebox{1mm}{\scriptsize{SMAPE}}}
        \psfrag{s2}{\hspace{-5mm}\raisebox{1mm}{\scriptsize{MASE}}}
        \psfrag{s3}{\hspace{-5mm}\raisebox{1mm}{\scriptsize{RMSE}}}
         \psfrag{V}{\hspace{0mm}\raisebox{-0.3mm}{\scalebox{0.6}{V-SLSTM}}}
        \psfrag{G}{\hspace{0mm}\raisebox{-0.5mm}{\scalebox{0.6}{G-SLSTM}}}
         \psfrag{v}{\hspace{0mm}\raisebox{-0.3mm}{\scalebox{0.6}{V-LSTM}}}
        \psfrag{g}{\hspace{0mm}\raisebox{-0.5mm}{\scalebox{0.6}{G-LSTM}}}
  \psfrag{1}{\hspace{0mm}\raisebox{-2mm}{\footnotesize{1}}}
            \psfrag{4}{\hspace{0mm}\raisebox{-2mm}{\footnotesize{4}}}
            \psfrag{7}{\hspace{0mm}\raisebox{-2mm}{\footnotesize{7}}}
          \psfrag{10}{\hspace{0mm}\raisebox{0mm}{\footnotesize{10}}}
          \psfrag{23}{\hspace{-2mm}\raisebox{0mm}{\footnotesize{23}}}
          \psfrag{36}{\hspace{-2mm}\raisebox{0mm}{\footnotesize{36}}}
          \psfrag{-0.3}{\hspace{-1mm}\raisebox{0mm}{\footnotesize{-0.3}}}
          \psfrag{1.1}{\hspace{-2mm}\raisebox{0mm}{\footnotesize{1.1}}}
          \psfrag{2.5}{\hspace{-2mm}\raisebox{0mm}{\footnotesize{2.5}}}
          \psfrag{0}{\hspace{0mm}\raisebox{0mm}{\footnotesize{0}}}
          \psfrag{11}{\hspace{-1mm}\raisebox{0mm}{\footnotesize{11}}}
          \psfrag{24}{\hspace{-1mm}\raisebox{0mm}{\footnotesize{24}}}

        \includegraphics[height = 1.1in, width = 0.3\linewidth]{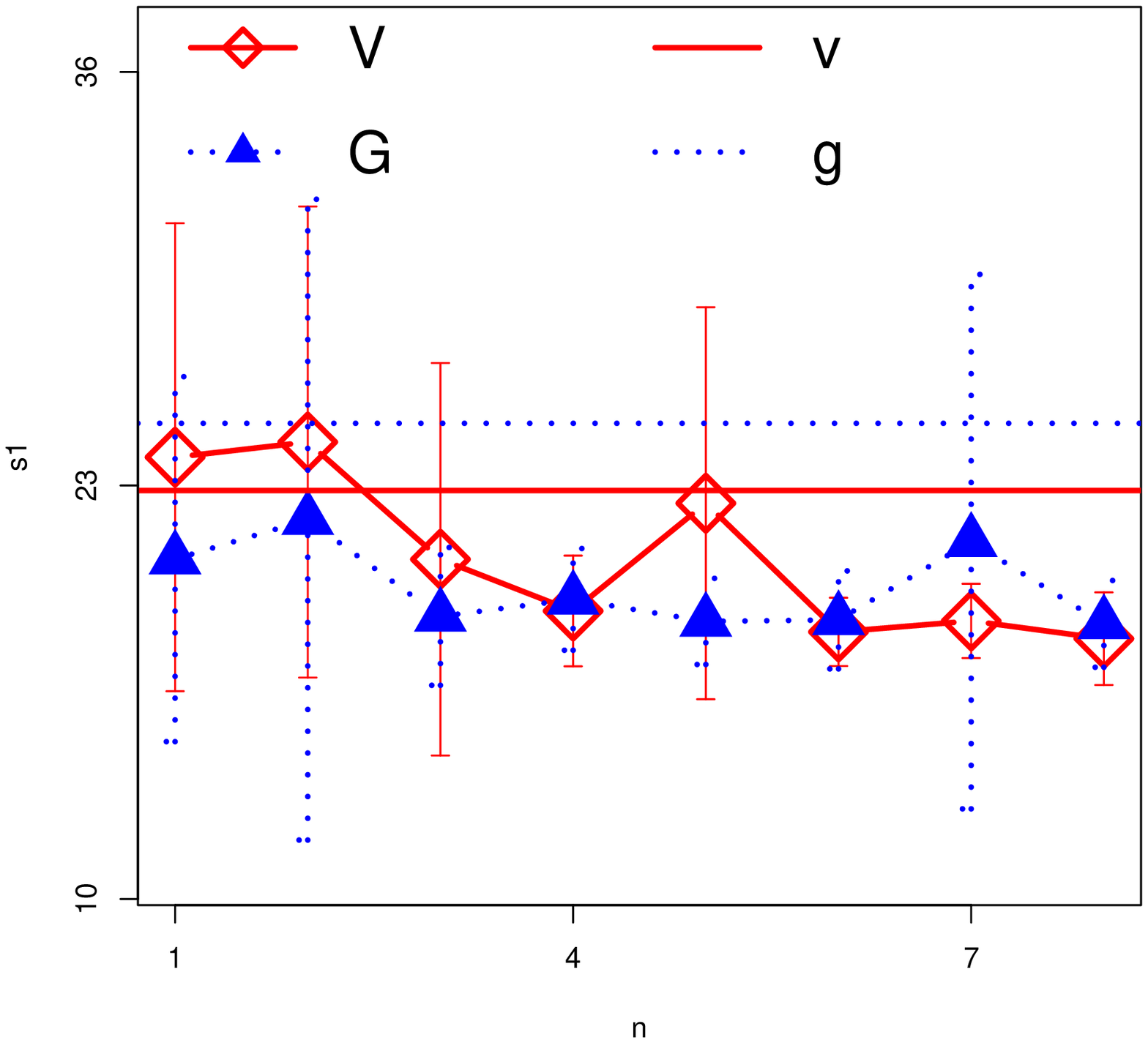}
        \includegraphics[height = 1.1in,width = 0.3\linewidth]{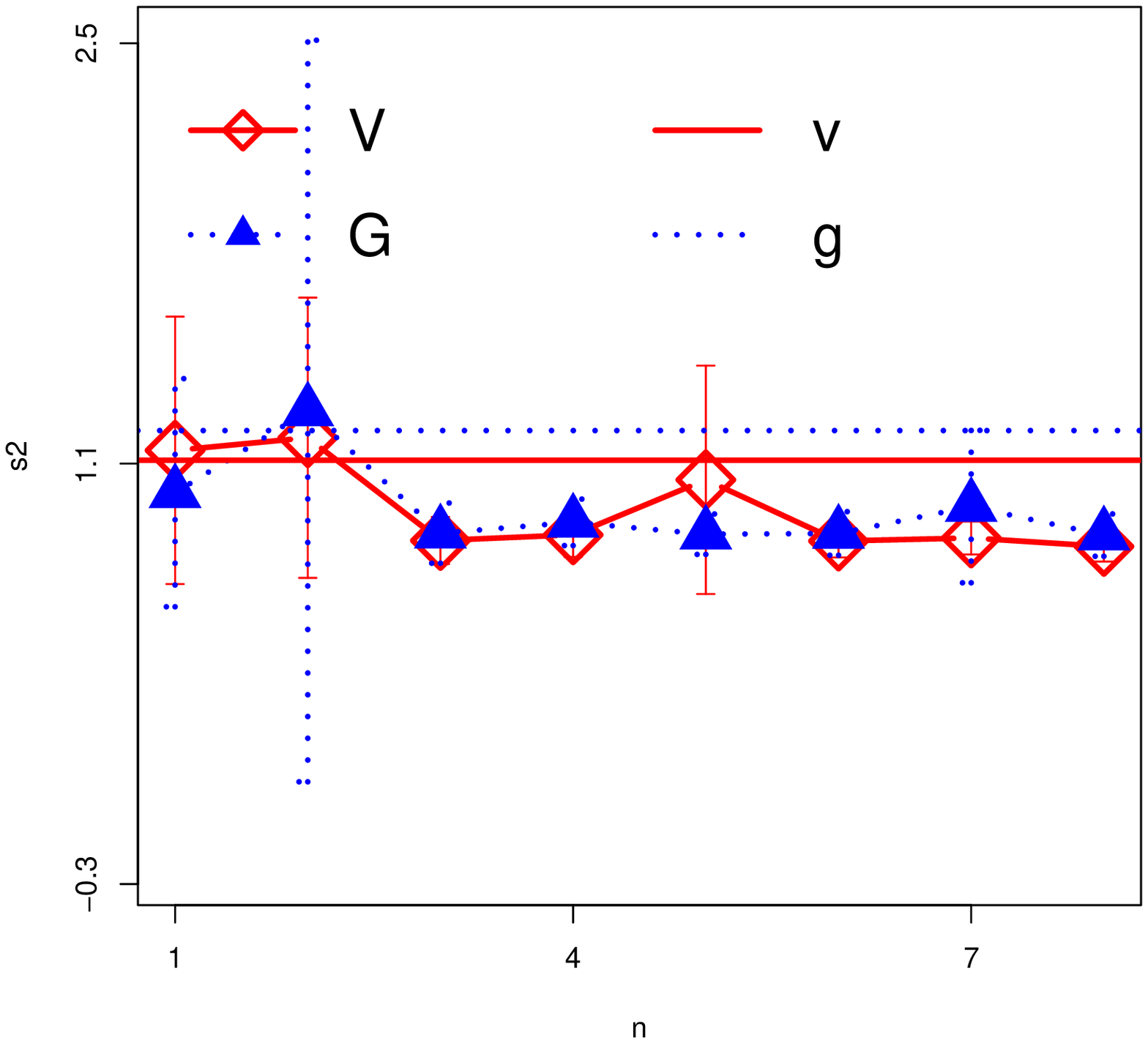}  
         \includegraphics[height = 1.1in, width = 0.3\linewidth]{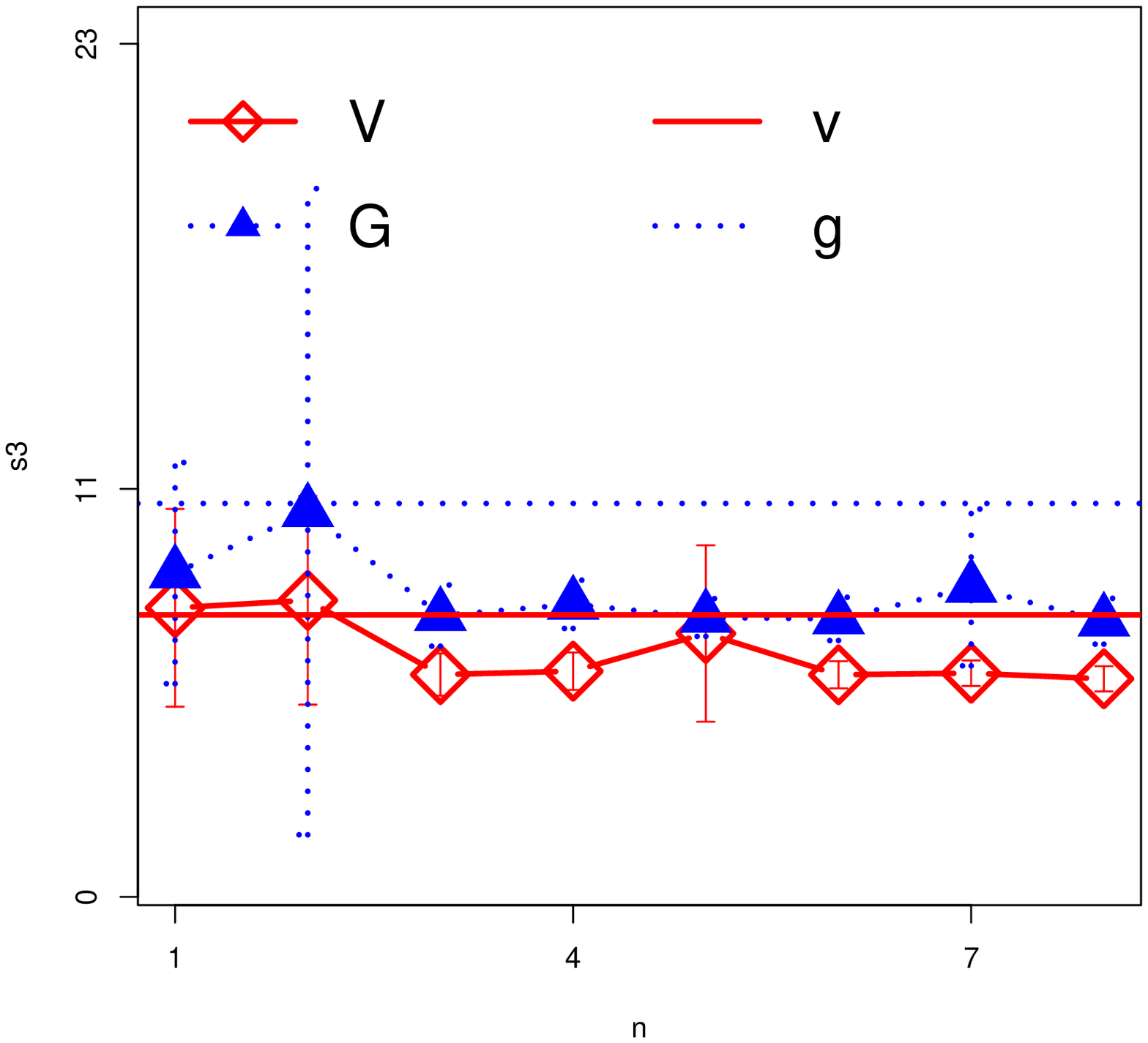}
        \vspace{1mm}
         \caption{Bengaluru demand data}
        \end{subfigure}%
        \vskip\baselineskip
        \begin{subfigure}{\textwidth}
        \centering
    \psfrag{n}{\hspace{-11mm} \raisebox{-3mm}{\footnotesize{Spatial features}}}
        \psfrag{s1}{\hspace{-5mm}\raisebox{1mm}{\scriptsize{SMAPE}}}
        \psfrag{s2}{\hspace{-5mm}\raisebox{1mm}{\scriptsize{MASE}}}
        \psfrag{s3}{\hspace{-5mm}\raisebox{1mm}{\scriptsize{RMSE}}}
         \psfrag{V}{\hspace{0mm}\raisebox{-0.2mm}{\scalebox{0.6}{V-SLSTM}}}
        \psfrag{G}{\hspace{0mm}\raisebox{-0.5mm}{\scalebox{0.6}{G-SLSTM}}}
         \psfrag{v}{\hspace{0mm}\raisebox{-0.2mm}{\scalebox{0.6}{V-LSTM}}}
        \psfrag{g}{\hspace{0mm}\raisebox{-0.5mm}{\scalebox{0.6}{G-LSTM}}}
  \psfrag{1}{\hspace{0mm}\raisebox{-2mm}{\footnotesize{1}}}
            \psfrag{4}{\hspace{0mm}\raisebox{-2mm}{\footnotesize{4}}}
            \psfrag{7}{\hspace{0mm}\raisebox{-2mm}{\footnotesize{7}}}
          \psfrag{-100}{\hspace{-1mm}\raisebox{0mm}{\footnotesize{-100}}}
          \psfrag{200}{\hspace{-2mm}\raisebox{0mm}{\footnotesize{200}}}
          \psfrag{500}{\hspace{-3mm}\raisebox{0mm}{\footnotesize{500}}}
          \psfrag{-0.6}{\hspace{-1mm}\raisebox{0mm}{\footnotesize{-0.6}}}
          \psfrag{2.6}{\hspace{-1mm}\raisebox{0mm}{\footnotesize{2.6}}}
          \psfrag{5.7}{\hspace{-2mm}\raisebox{0mm}{\footnotesize{5.7}}}
          \psfrag{0}{\hspace{0mm}\raisebox{0mm}{\footnotesize{0}}}
          \psfrag{18}{\hspace{0mm}\raisebox{0mm}{\footnotesize{18}}}
          \psfrag{37}{\hspace{-2mm}\raisebox{0mm}{\footnotesize{37}}}

        \includegraphics[height = 1.1in, width = 0.3\linewidth]{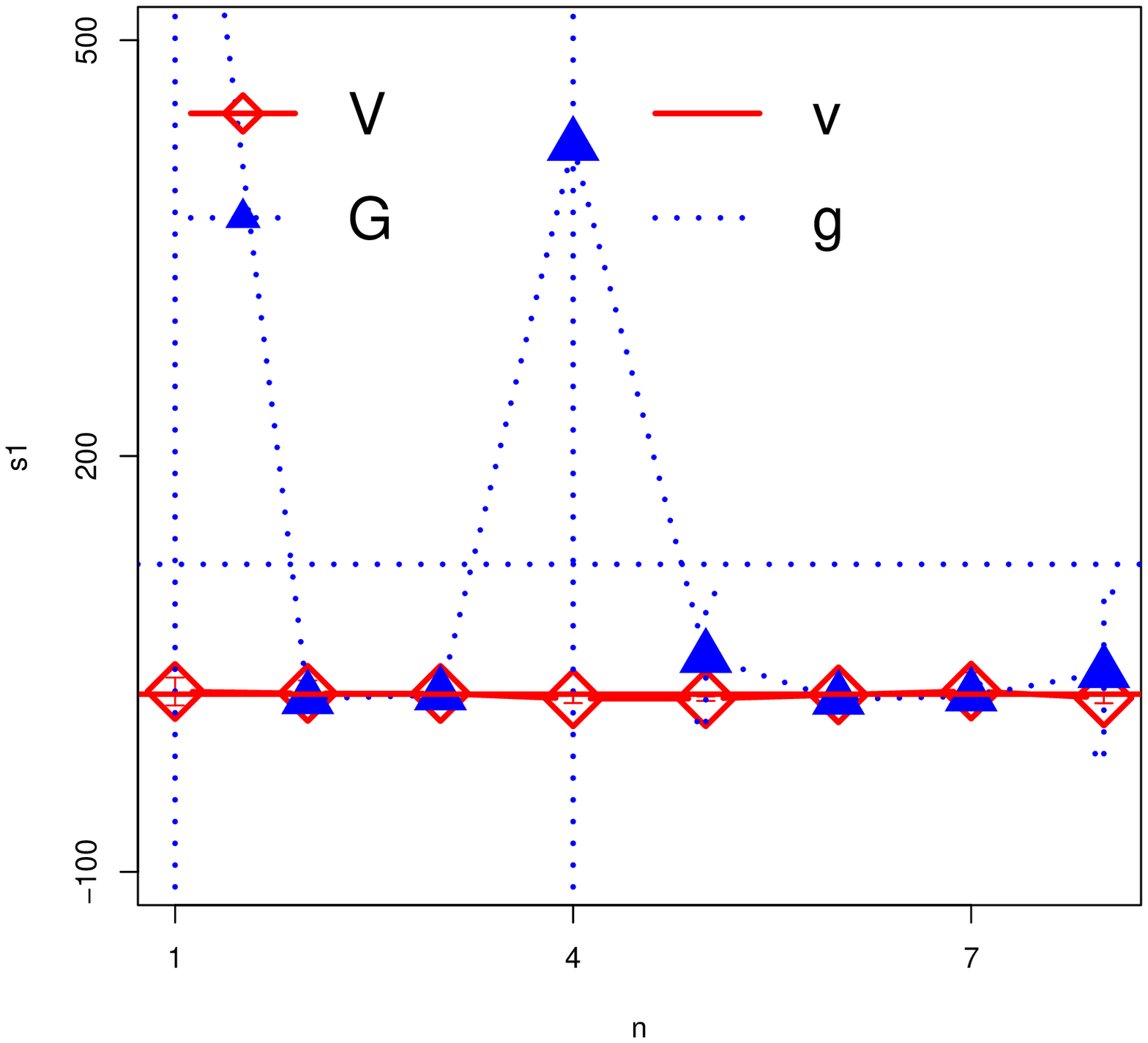}
         \includegraphics[height = 1.1in, width = 0.3\linewidth]{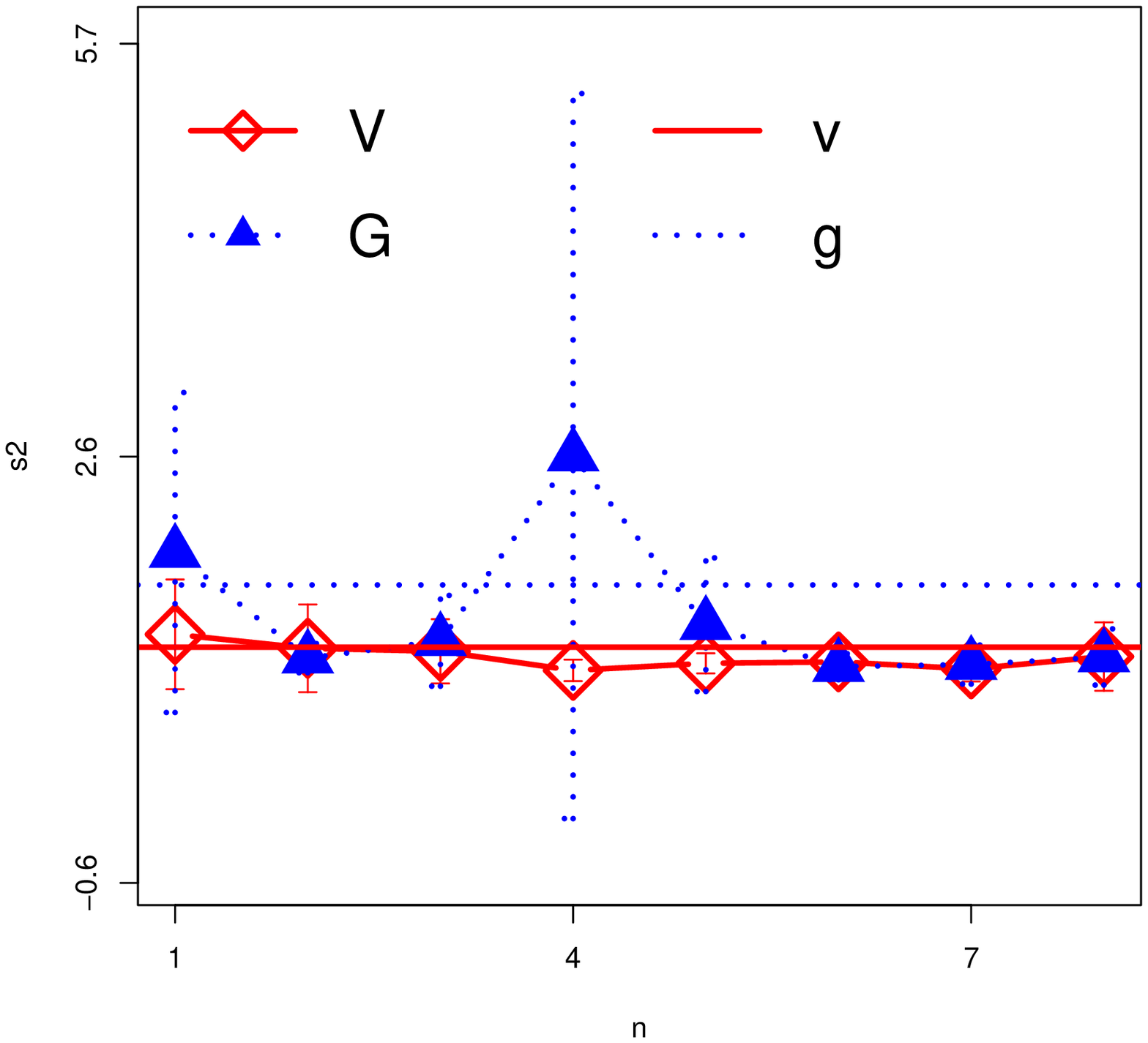}
        \includegraphics[height = 1.1in, width = 0.3\linewidth]{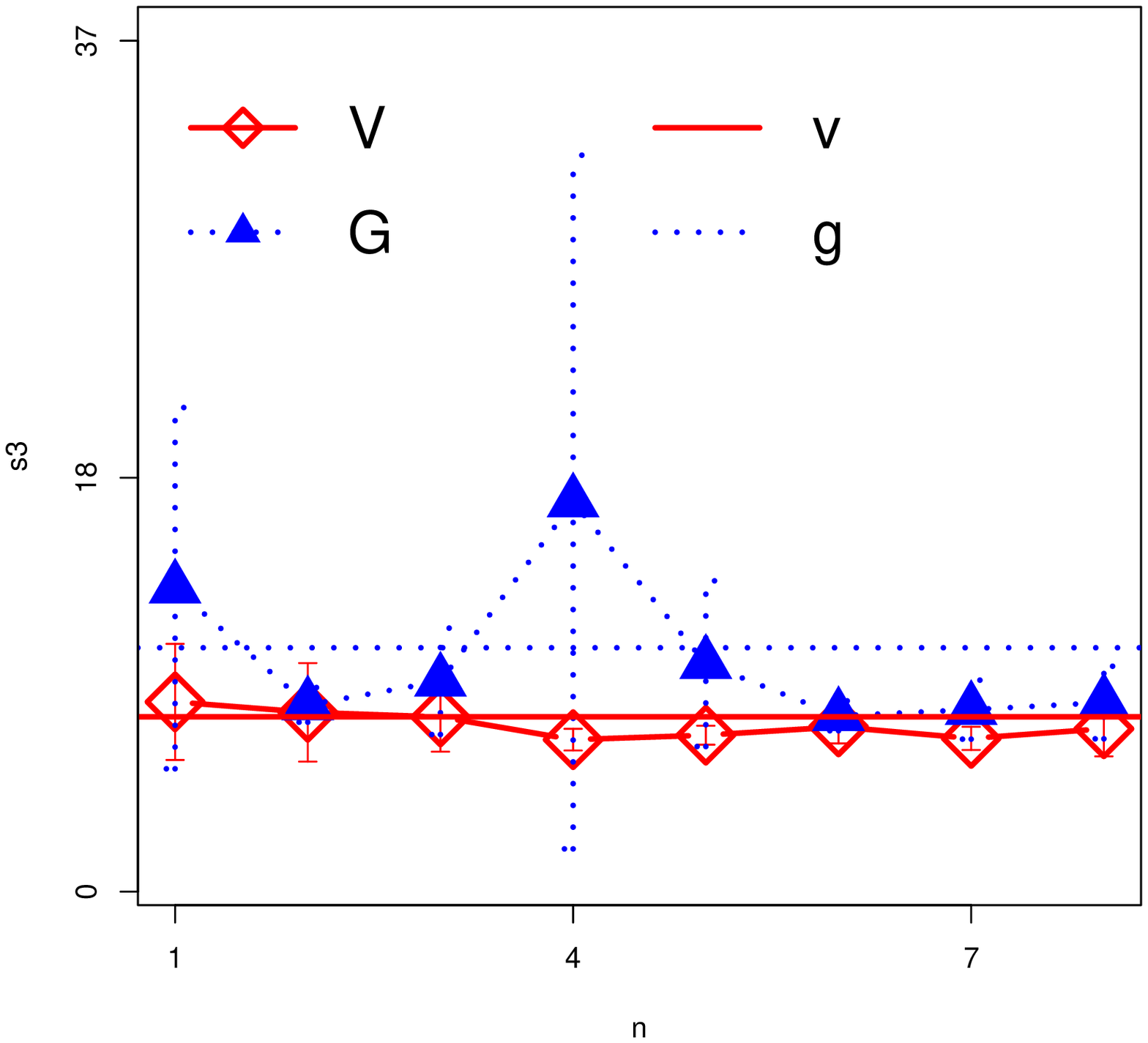}
        \vspace{1mm}
         \caption{Bengaluru supply data}
        \end{subfigure}%
        \caption{ Plots showing performance of LSTM and Spatial LSTM (with variable number of features) on the demand and supply data sets. V-SLSTM, G-SLSTM, V-LSTM, and G-LSTM denote Voronoi and Geohash based LSTM, with and without spatial features.} 
        \label{resultsfig}
\end{figure*}

From the numerical results presented in the Table \ref{results}, the models based on Voronoi features exhibit better prediction accuracy than the models based on Geohash features, with all three metrics considered. This behavior appears to be consistent across scenarios with and without spatial information. However, the percentage improvement in accuracy with spatial information is higher in Geohash based models. When we varied the number of input spatial features, both Voronoi and Geohash based models achieved best prediction accuracy with 8 spatial features for demand data. For the supply data set, the best configuration turned out to be LSTM with 4 spatial features for Voronoi and 6 spatial features for Geohash. In other words, the best Voronoi based LSTM setting was achieved with same or less number of spatial features than the best Geohash based LSTM setting. We observed more variability in the predictions of Geohash based models than of Voronoi based models over multiple runs. As can be seen from Figure \ref{resultsfig}, the standard deviations of the errors from Geohash based models are higher than the standard deviations of errors observed from Voronoi based models. For our tested scenarios, we observe that Voronoi based models produce more stable results across multiple runs. 

\begin{table}[h!]
    \caption{The average error levels of LSTM with and without spatial features, for Bengaluru and New York data sets. We use LSTM with one hidden layer, Linear activation, and vary learning rate of Adam \cite{kingma2014adam}, keeping other parameters of Adam at their default values. The numbers in brackets denote the number of spatial features that resulted in the best model performance.}
    \label{results}
    \begin{center}
    \scalebox{0.79}{
    {\renewcommand{\arraystretch}{1.4}
    \begin{tabular}{llllllll}
\hline
\multirow{3}{*}{\quad \quad Data Set} & \multirow{3}{*}{\quad \quad HyperOpt Parameters}               & \multicolumn{3}{c}{LSTM} & \multicolumn{3}{c}{Spatial LSTM} \\ \cline{3-8} 
                         &                                                                                              & SMAPE   & MASE   & RMSE & SMAPE  & MASE    & RMSE    \\ \cline{3-8} 
                         &                                                                                              & \multicolumn{6}{c}{Voronoi Tessellation}                     \\ \hline
Bengaluru Demand     & n = {[}10{]}, D = 0.137, Adam ($\text{10}^{-1}$) & 22.8    & 1.11   & 7.60   & 18.1 (8)  & 0.82 (8)  & 5.87 (8)  \\ \hline
Bengaluru Supply        & n = {[}20{]}, D = 0.368, Adam ($\text{10}^{-1}$) & 28.2    & 1.16   & 7.76   & 24.9 (4)  & 0.99 (4)  & 6.60 (4)  \\ \hline
New York Demand          & n = {[}20{]}, D = 0.217, Adam ($\text{10}^{-2}$) & 27.1    & 1.08   & 9.70  & 21.2 (6)  & {0.83 (6)}  & {7.83 (6)}  \\ \hline
\multicolumn{2}{c}{}                                                                                                  & \multicolumn{6}{c}{Geohash Tessellation}                     \\ \hline
Bengaluru Demand        & n = {[}10{]}, D = 0.236, Adam ($\text{10}^{-3}$) & 24.9    & 1.21   & 10.6   & 18.6 (8)  & 0.86 (8)  & 7.43 (8)  \\ \hline
Bengaluru Supply        & n = {[}20{]}, D = 0.325, Adam ($\text{10}^{-1}$) & 121.9    & 1.63   & 12.2   & 24.4 (6)  & 1.02 (6)  & 7.64 (6)  \\ \hline
New York Demand          & n = {[}20{]}, D = 0.173, Adam ($\text{10}^{-1}$)  & 15.1    & 0.86   & 57.5   & 14.9 (4)  & 0.85 (4)  & 56.1 (4)  \\ \hline
    \end{tabular}}}
    \end{center}
\end{table}

\subsection{New York demand data} \label{nyc}
The publicly available New York yellow taxi data set \cite{nyc} contains GPS traces of a street hailing yellow taxi service.  We extract the pick up locations and time stamps from the period of January-February 2016, for analysis. As explained in Section \ref{methodology}, we aggregate the data in Voronoi cells and geohashes. From Table \ref{results}, we see that while Voronoi based models outperform Geohash based models with RMSE and MASE, Geohash models have lower SMAPE than Voronoi models. Voronoi models incorporated more spatial information and exhibited lower variability in the prediction errors than the Geohash models. The non-uniform distribution of demand in New York city appears to be a plausible reason behind the high SMAPE for Voronoi models. The non-uniformity in distribution results in many Voronoi cells with lower demand volume as compared to their corresponding geohashes. This drives the scale dependent SMAPE high in several Voronoi cells. In fact, these results corroborate the findings in \cite{davis2018taxi}, where the authors used classical time-series techniques to model taxi demand data sets. These comparisons suggest that the choice of the tessellation strategy might depend on the performance metric and data set employed for the study. Nevertheless, the performance enhancement observed with Voronoi based New York demand model, using two commonly used performance metrics, is significant and merits attention.
 
\subsection{Comparison with Convolutional LSTM}
Convolutional LSTM (convLSTM) \cite{xingjian2015convolutional} is an extension of the fully connected LSTM network with convolutional structures in both input-to-state and state-to-state transitions. Several variants of convLSTM have been proposed for taxi travel forecasting, but the feature extraction remains limited to a fixed grid city structure. This is because the CNNs are basically designed to work on regular grids, and cannot process graphs.  We proceed to perform a comparison between LSTM with Geohash features, convLSTM with input frames obtained from Geohash grids, and LSTM with Voronoi features. The analysis was performed on the Bengaluru demand and supply data sets. We use a 2D convLSTM layer with input shape (740, 1440, 3, 3, 1) feeding into a 3D CNN layer, with frames of size 3 $\times$ 3, 1 channel, and 8 filters. MSE is employed as the loss function with Adam optimizer and the simulations ran for 500 training epochs. Predictions from the Geohash convLSTM model resulted in a MASE of 0.55 and 0.42, RMSE of 2.8 and 1.94, and SMAPE of 50.7 and 74.2, for demand and supply data respectively. On referring Table \ref{results} for results on LSTM, we see that the convLSTM, as expected, performed better than LSTM on MASE and RMSE. But, with SMAPE as the metric, we see that a Voronoi LSTM has a lower error compared to a Geohash convLSTM network. A graph based convLSTM that can work with Voronoi features is promising in this scenario. 

\section{Concluding remarks} \label{conclusions}
Taxi demand-supply forecasting is an integral part of Intelligent Transportation Systems. In this paper, we use the Long Short-Term Memory (LSTM) network to compare the variable-sized polygon based Voronoi tessellation and the fixed-sized grid based Geohash tessellation on demand-supply data sets from the cities of Bengaluru, India and New York, USA. We found that the LSTM network based on spatial features obtained from Voronoi tessellation has a higher prediction accuracy than the LSTM network based on spatial features obtained from Geohash tessellation. With Voronoi features, we noticed less variability in the results obtained across multiple runs.  

Comparisons with conventional Convolutional LSTM networks highlight the need to explore graph based Convolutional LSTM networks. The results suggest that a graph based Convolutional LSTM might outperform the conventional grid based Convolutional LSTM, since the Voronoi LSTM already has a superior performance over Geohash LSTM. In fact, there have been exciting developments in the area of graph CNNs \cite{nikolentzos2018kernel, defferrard2016convolutional}, which may be extended to graph convLSTMs.

The findings in this paper recommend exploration and selection of a suitable tessellation strategy prior to fitting a neural network model. In currently employed LSTM networks, the input features are commonly extracted from a city space that is divided into fixed-sized grids. An appropriate tessellation strategy might reduce the complexity of the neural network to be built. For instance, in our study, we see that a LSTM with Voronoi based features had a better SMAPE metric than a more sophisticated Convolutional LSTM with Geohash based features.



\end{document}